\documentclass[sn-mathphys,Numbered]{sn-jnl}% Math and Physical Sciences Reference Style
\usepackage{graphicx}%
\usepackage{multirow}%
\usepackage{amsmath,amssymb,amsfonts}%
\usepackage{amsthm}%
\usepackage{mathrsfs}%
\usepackage[title]{appendix}%
\usepackage{xcolor}%
\usepackage{textcomp}%
\usepackage{manyfoot}%
\usepackage{booktabs}%
\usepackage{algorithm}%
\usepackage{algorithmicx}%
\usepackage{algpseudocode}%
\usepackage{listings}%
\theoremstyle{thmstyleone}%
\theoremstyle{thmstyletwo}%

\theoremstyle{thmstylethree}%

\begin{document}

\title{Graph Neural Networks for Antisocial Behavior Detection on Twitter}


\author*[1]{\fnm{Martina} \sur{Toshevska}}\email{martina.toshevska@finki.ukim.mk}

\author[1]{\fnm{Slobodan} \sur{Kalajdziski}}\email{slobodan.kalajdziski@finki.ukim.mk}

\author[1]{\fnm{Sonja} \sur{Gievska}}\email{sonja.gievska@finki.ukim.mk}

\affil[1]{\orgdiv{Faculty of Computer Science and Engineering}, \orgname{Ss. Cyril and Methodius University}, \orgaddress{\city{Skopje}, \country{North Macedonia}}}

\abstract{Social media resurgence of antisocial behavior has exerted a downward spiral on stereotypical beliefs, and hateful comments towards individuals and social groups, as well as false or distorted news. The advances in graph neural networks employed on massive quantities of graph-structured data raise high hopes for the future of mediating communication on social media platforms. An approach based on graph convolutional data was employed to better capture the dependencies between the heterogeneous types of data. 

Utilizing past and present experiences on the topic, we proposed and evaluated a graph-based approach for antisocial behavior detection, with general applicability that is both language- and context-independent. In this research, we carried out an experimental validation of our graph-based approach on several PAN datasets provided as part of their shared tasks, that enable the discussion of the results obtained by the proposed solution.}




\maketitle

\section{Introduction}
 
With the rise of social media platforms, interpersonal communication has become easier and more frequent. However, antisocial behavior has also experienced an increase in various forms such as stereotypical or hateful comments toward individuals or social groups, false or distorted news, aggression, violence, etc. Although it could be beneficial for the author in terms of reaching more audiences or getting more views, likes, etc., it can be harmful to the target. Being able to detect online antisocial behavior could be a significant asset for social media platforms that enable them to perform actions to prevent it.

Graph Neural Networks (GNNs) are deep learning-based models that operate on graph structures. GNNs learn embedding representation for each node in the graph. Edge embeddings and graph embeddings can be created with the aggregation of node embeddings. GNNs perform two operations on the node embeddings obtained by the previous layer and the adjacency matrix of the graph~\cite{kipf2017semi, defferrard2016convolutional, hamilton2017inductive, wu2021graph}. The first operation is graph filtering which computes node embeddings, while the second is graph pooling which generates a smaller graph with fewer nodes and its corresponding new node embeddings. There is a variety of GNN models that implement various graph filtering functions.

In the past few years, GNNs have gained interest in the Natural Language Processing (NLP) field for text classification~\cite{yao2019graph, lu2020vgcn}. The traditional models based on recurrent neural networks (RNNs), convolutional neural networks (CNNs), and/or transformers capture contextual (local) information within a sentence. On the other hand, graph-based approaches capture global information about the vocabulary of a language~\cite{lu2020vgcn}. Since the text data does not naturally have a graph structure, the crucial and most important part is to represent the text as a graph. Early approaches are focused on constructing text graphs composed of word nodes and documents nodes~\cite{yao2019graph}, while more recent approaches demonstrate that augmenting with additional information such as part of speech (POS) tags, named entities, and transformer-based word/sentence embeddings is beneficial.

In this paper, we evaluate the performance of several graph neural networks on the problem of detecting fake news and hate speech spreaders on Twitter\footnote{The code for this research is available at: https://github.com/mtoshevska/Antisocial-Behavior-on-Twitter}. We define the problem as a node classification problem. We have created heterogeneous graphs using the datasets provided by a series of shared tasks on digital text forensics and stylometry (PAN) and we have trained several graph neural network models to classify user nodes. For comparison we have evaluated the proposed models on two additional tasks i.e. irony/stereotype spreaders on Twitter and sentiment classification on Yelp reviews. The rest of the paper is organized as follows. In Section~\ref{sec:related_work}, a brief introduction to GNN approaches to text classification problems is presented. The datasets are described in detail in Section~\ref{sec:datasets}. Section~\ref{sec:baseline_models} presents the baseline models. The heterogeneous graph creation process is presented in Section~\ref{sec:graph_creation}, while Section~\ref{sec:gnn_models} describes graph neural network models. The results are presented in Section~\ref{sec:results}. Section~\ref{sec:conclusion} concludes the paper.

\section{Related Work}
\label{sec:related_work}

Graph-based approaches have been evaluated for many text classification tasks. TextGCN~\cite{yao2019graph} operates on a heterogeneous graph created from text data representing words and documents as nodes, and relations between them as edges. Two-layer graph convolutional network (GCN) is applied on the heterogeneous text graph to allow indirect message passing between document nodes. TextGCN significantly outperforms baseline RNN-/CNN-based models on several benchmark datasets for sentiment classification, newsgroup classification, medical abstract classification, etc. The heterogeneous graph in our study was created following the TextGCN process of graph creation.

VGCN-BERT~\cite{lu2020vgcn} augments a BERT-based text classification model with graph embeddings to include global information about the vocabulary. A vocabulary graph has been constructed using normalized point-wise mutual information (NPMI). Vocabulary GCN (VGCN) has been applied to the vocabulary graph to create a graph embedding for the sentence. VGCN captures the part of the graph relevant to the input and then performs 2 layers of convolution, combining words from the input sentence with their related words in the vocabulary graph. To obtain the final class prediction, multiple layers of attention mechanism have been applied to the concatenated representation of the input text created with BERT and graph embeddings created with VGCN. VGCN-BERT has been evaluated on multiple text classification tasks including sentiment classification, hate speech detection, etc. In~\cite{lin2021bertgcn}, a heterogeneous graph has been constructed following TextGCN~\cite{yao2019graph}, but a BERT/RoBERTa model has been used to obtain embeddings for the initial representation of the document nodes. The proposed model, BertGCN, has been optimized jointly with an auxiliary classifier that directly operates on BERT embeddings because it led to faster convergence and better performances. BertGCN parameters have been initialized with parameters of a pre-trained BERT model on the target dataset to speed up the training. Compared with the traditional BERT/RoBERTa models, the BertGCN yielded better performances. BertGCN has been evaluated on the same benchmark datasets as TextGCN. The performance gains obtained by BertGCN were higher for datasets containing longer sentences that enable capturing longer-term dependencies. Node representation in our study follows the BertGCN idea of document representation. We have utilized a BERT-based model to create an embedding for the initial representation of each tweet.

PAN\footnote{https://pan.webis.de/, last visited: 25.02.2023} is a series of scientific events and shared tasks on digital text forensics and stylometry. There is a series of author profiling shared tasks that each year are focused on a different topic. In the past three years, they were focused on antisocial behavior detection on Twitter. The participants have used a wide variety of models starting from traditional machine learning models to Transformer-based architectures~\cite{rangel2020overview, rangel2021profiling, reynier2022profiling}. Most of the participants have used traditional machine learning approaches with various features such as n-grams, term frequency-inverse document frequency (TF-IDF), lexicons, word embeddings, sentence embeddings, etc. A few of the participants in 2020~\cite{rangel2020overview} have created deep learning models such as multi-layer perceptron (MLP), CNNs, and RNNs. In the 2021 shared task~\cite{rangel2021profiling}, one of the participants built a BERT-based model with additional linear layers; and in 2022~\cite{reynier2022profiling}, a graph convolutional neural network was first implemented by one of the participants. The best performing model for the task of fake news spreaders detection was a Logistic Regression model trained with n-gram features, as well as some statistic-based features from the tweets such as average length or lexical diversity~\cite{buda2020ensemble}. The best performing model for the task of hate speech spreaders detection was a CNN model that used 100-dimensional word embedding vectors~\cite{siino2021detection}. The best performing model for the task of detecting irony and stereotype spreaders was a CNN model with BERT-based tweet features~\cite{yu2022bert}. In our experiments, we have used datasets provided by the PAN shared tasks. Since there was only one participant utilizing GNNs for the shared tasks, we aim to investigate in detail the performance of GNNs on these datasets.

\section{Datasets}
\label{sec:datasets}

The datasets used for the experiments are provided by PAN for the Author Profiling shared tasks for the years 2020 (Profiling fake news spreaders on Twitter) and 2021 (Profiling hate speech spreaders on Twitter). The dataset for the 2022 shared task (Profiling irony and stereotype spreaders on Twitter - IROSTEREO) and a dataset for sentiment classification were also used to evaluate and compare the performances of the proposed models with more data.

\subsection{Profiling Fake News Spreaders on Twitter}

The training set provided in the Profiling Fake News Spreaders on Twitter\footnote{https://pan.webis.de/clef20/pan20-web/author-profiling.html, last visited: 25.02.2023} shared task is composed of 300 Twitter users with 100 tweets per user. Each user is labeled as either user posting tweets that contain fake news (1) or a user posting tweets that do not contain fake news (0). We have randomly chosen 80\% of the users for training and 20\% for validation in a way that the proportion of users in each class is retained. The testing set is composed of 200 Twitter users with 100 tweets per user.

\subsection{Profiling Hate Speech Spreaders on Twitter}

The training set provided in the Profiling Hate Speech Spreaders on Twitter\footnote{https://pan.webis.de/clef21/pan21-web/author-profiling.html, last visited: 25.02.2023} is composed of 200 Twitter users with 200 tweets per user. Each user is labeled as either user posting tweets that contain hate speech (1) or a user posting tweets that do not contain hate speech (0). Because the testing set was not available, we randomly split the users in the training set into subsets for training (80\%), validation (10\%), and testing (10\%).

\subsection{Profiling Irony and Stereotype Spreaders on Twitter - IROSTEREO}

The training set provided in the Profiling Irony and Stereotype Spreaders on Twitter\footnote{https://pan.webis.de/clef22/pan22-web/author-profiling.html, last visited: 25.02.2023} shared task is composed of 420 Twitter users with 200 tweets per user. Each user is labeled as either user posting ironic tweets (I) or a user not posting ironic tweets (NI). The testing set is composed of 180 Twitter users with 200 tweets per user. Although the testing set was available, the ground truth labels were not provided. We have created a training, validation, and testing subset by randomly choosing 80\%, 10\%, and 10\% of the users, respectively.

In this shared task, another dataset for stereotype stance detection was provided. It contains the users that are labeled as users that are posting ironic tweets. Each user is labeled as either user posting ironic tweets with stereotypes in favor of the target (INFAVOR) or a user posting ironic tweets with stereotypes against the target (AGAINST). The training set is composed of 140 Twitter users, while the testing set is composed of 60 Twitter users. The number of tweets per user is 200. For this dataset, the testing set was available, but ground truth labels were not. We have created a training, validation, and testing subset by randomly choosing 80\%, 10\%, and 10\% of the users, respectively.

\subsection{Yelp Open Dataset}

Yelp\footnote{https://www.yelp.com/dataset, last visited: 25.02.2023} dataset is a collection of 8.6 million business reviews that are rated with a 5-star rating system. We have created labels for the reviews according to the rating as follows. If the rating is less than or equal to 3 the review is labeled as negative, and as positive if the rating is greater than 3. The dataset was filtered in a way that the number of reviews per user is similar to the number of tweets per user in the previous datasets and the review length is similar to the tweet length. It has been filtered first by the number of reviews per user and then by the length of the reviews. We kept only the reviews written by users with a number of written reviews in the range from 50 to 200 with a length in the range from 15 to 60. Using the remaining reviews, we have created a training, validation, and testing subset by randomly choosing 80\%, 10\%, and 10\% of the users, respectively.

\section{Baseline Models}
\label{sec:baseline_models}

Following the success of the Transformer architectures for many natural language processing tasks and to compare the performance of the graph neural network models, we have trained three Transformer-based models: DistilBERT~\cite{sanh2019distilbert}, RoBERTa~\cite{liu2019roberta}, and DistilRoBERTa~\cite{sajjad2020effect}. DistilBERT learns an approximate version of BERT using a knowledge distillation technique~\cite{buciluǎ2006model, hinton2015distilling}. With only one-half of the layers of the original version of the BERT model, the number of parameters is reduced by 40\%. DistilBERT is designed to be smaller and faster than BERT, while still retaining much of its accuracy. RoBERTa follows the original BERT architecture but has been trained with a different training procedure and on a larger corpus of text. It has been trained with dynamic masking where the masking pattern is generated every time a sequence is fed to the model, as opposed to static masking in the original BERT implementation where the same training mask was used. RoBERTa has been trained without the next sentence prediction objective, with bigger batches over more data and longer sequences. DistilRoBERTa is a combination of the former two models. It learns an approximate version of the RoBERTa model following the same training procedure as in DistilBERT.

Since the goal is to classify users based on the tweets they have posted, we have concatenated all tweets of a particular user into one representation. We have used PyTorch implementation of these models available in the Huggingface Transformers library\footnote{https://huggingface.co/docs/transformers/index, last visited: 25.02.2023}. We initialized the weights with the pre-trained \textit{distilbert-base-uncased}, \textit{roberta-base}, and \textit{distilroberta-base} weights for DistilBERT, RoBERTa, and DistilRoBERTa models, respectively. All models have been trained with AdamW optimizer, binary cross-entropy loss, and batch size 16. For the other hyperparameters, we have performed a hyperparameter search among a set of possible values. The optimal hyperparameters for each model and each dataset are summarized in Table~\ref{tab:transformers_hyperparameters}.

\begin{table}[]
  \caption{Optimal hyperparameters for Transformer models.}
  \label{tab:transformers_hyperparameters}
  \begin{tabular}{r|c|c|c}
    \hline
    \multicolumn{4}{c}{\textbf{Fake News}} \\ \hline
    & \textbf{Learning Rate} & \textbf{Weight Decay} & \textbf{Epochs} \\ \hline
    \textbf{DistilBERT} & 0.00001 & 0.005 & 100 \\
    \textbf{RoBERTa} & 0.00001 & 0.00005 & 250 \\
    \textbf{DistilRoBERTa} & 0.00001 & 0.005 & 250 \\ \hline \hline
    \multicolumn{4}{c}{\textbf{Hate Speech}} \\ \hline
    & \textbf{Learning Rate} & \textbf{Weight Decay} & \textbf{Epochs} \\ \hline
    \textbf{DistilBERT} & 0.00001 & 0.0005 & 250 \\
    \textbf{RoBERTa} & 0.00001 & 0.0005 & 250 \\
    \textbf{DistilRoBERTa} & 0.00001 & 0.0005 & 250 \\ \hline \hline
    \multicolumn{4}{c}{\textbf{Irony Stereotype}} \\ \hline
    & \textbf{Learning Rate} & \textbf{Weight Decay} & \textbf{Epochs} \\ \hline
    \textbf{DistilBERT} & 0.00001 & 0.005 & 500 \\
    \textbf{RoBERTa} & 0.00001 & 0.0005 & 100 \\
    \textbf{DistilRoBERTa} & 0.00001 & 0.005 & 100 \\ \hline \hline
    \multicolumn{4}{c}{\textbf{Stereotype Stance}} \\ \hline
    & \textbf{Learning Rate} & \textbf{Weight Decay} & \textbf{Epochs} \\ \hline
    \textbf{DistilBERT} & 0.00001 & 0.005 & 500 \\
    \textbf{RoBERTa} & 0.00001 & 0.005 & 100 \\
    \textbf{DistilRoBERTa} & 0.0001 & 0.005 & 100 \\ \hline \hline
    \multicolumn{4}{c}{\textbf{Yelp}} \\ \hline
    & \textbf{Learning Rate} & \textbf{Weight Decay} & \textbf{Epochs} \\ \hline
    \textbf{DistilBERT} & 0.00001 & 0.005 & 500 \\
    \textbf{RoBERTa} & 0.00001 & 0.00005 & 250 \\
    \textbf{DistilRoBERTa} & 0.00001 & 0.0005 & 250 \\
    \hline
  \end{tabular}
\end{table}

\section{Heterogeneous Graph Creation}
\label{sec:graph_creation}

\begin{figure}
  \centering
  \includegraphics[width=0.65\linewidth]{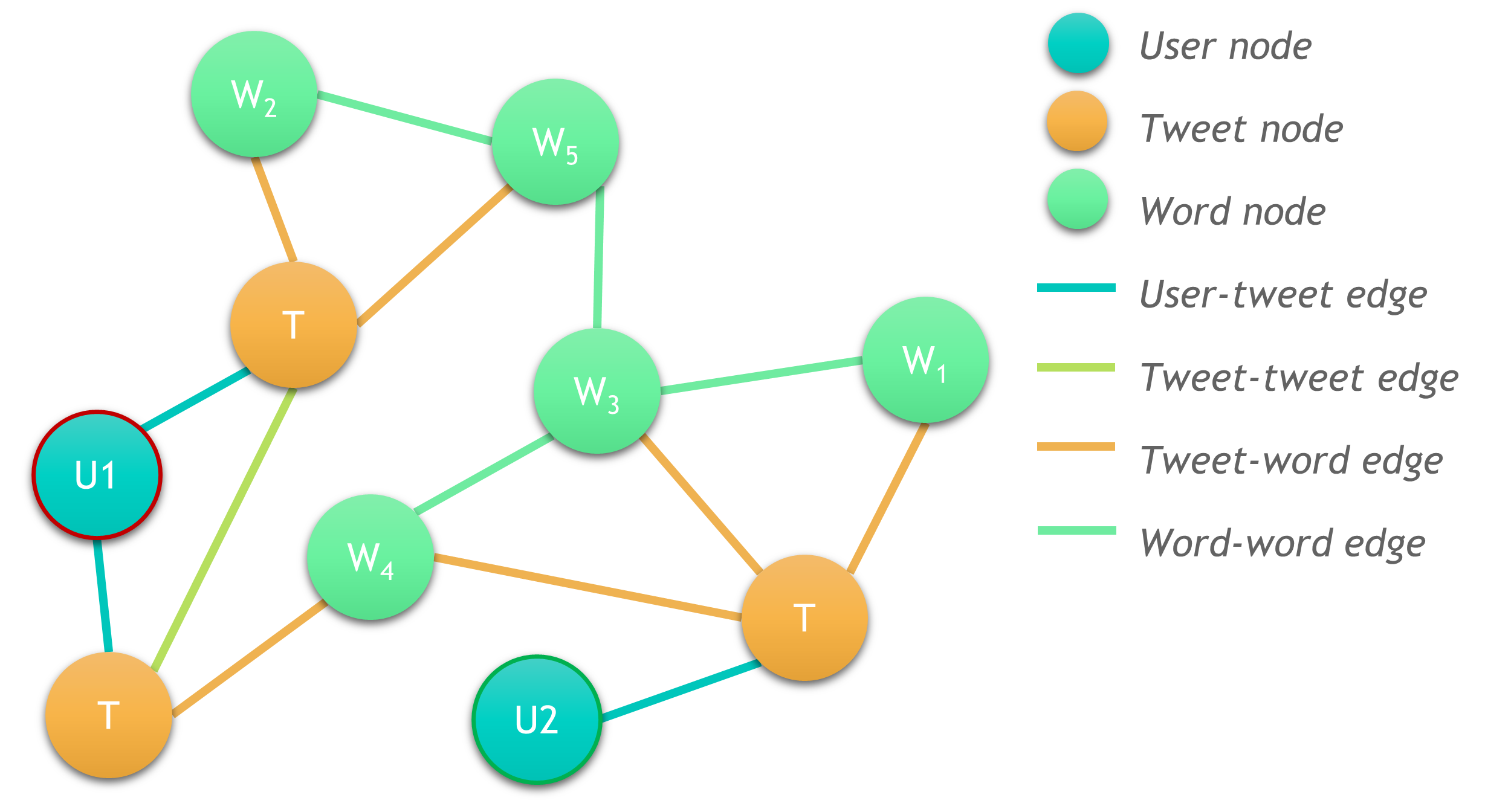}
  \caption{Simplified visualization of the heterogeneous graph. $U1$, $U2$ - user nodes. $P$ - tweet nodes. $W_1$-$W_5$ - word nodes. The user $U1$ represents a user from the first class (e.g. posting ironic tweets), while the user $U2$ represents a user from the second class (e.g. not posting ironic tweets).}
  \label{fig:heterogeneus_graph}
\end{figure}

We have created a heterogeneous graph dataset for classifying Twitter users, composed of three types of nodes: (1) user nodes, (2) tweet nodes, and (3) word nodes; and four types of edges: (1) user-tweet, (2) tweet-word, (3) word-word, and (4) tweet-tweet. The graph was created using all data in the subsets for training, validation, and testing. A simplified visualization of the graph is shown in Figure~\ref{fig:heterogeneus_graph}.

A vocabulary composed of the unique words in the dataset has been created. Special tokens representing user mentions, links, and hashtags have been added to the vocabulary. Rare words (words with less than 15 occurrences) have been removed and the remaining were used as word nodes.

Following the BertGCN~\cite{lin2021bertgcn} model, we utilize word and sentence embeddings to encode the nodes. Each word node is initialized with a word embedding of the corresponding word. We have used 200-dimensional GloVe~\cite{pennington2014glove} embedding vectors pre-trained on a Twitter dataset. The embeddings have been extracted using the Gensim library\footnote{https://radimrehurek.com/gensim/, last visited: 25.02.2023}.

Each tweet is represented as a node initialized with a 768-dimensional sentence embedding obtained by a pre-trained DistilRoBERTa~\cite{sajjad2020effect} model. The embeddings have been obtained using the Sentence-Transformers library\footnote{https://www.sbert.net/, last visited: 25.02.2023}. User nodes have been initialized via the embedding representation of their tweets. Pre-trained DistilRoBERTa embeddings have been obtained for each of the 200 tweets per user. The embeddings have been averaged along the 0-axis thus ending with a 768-dimensional representation for each user.

Word-word and tweet-word edges have been added following the graph creation process for the TextGCN model~\cite{yao2019graph}. Edges between a pair of word nodes are added if the PMI is greater than 0. PMI value has been set as a weight for word-word edges. Edges between words and tweets are added with the TF-IDF of the word in the tweet as a weight for the edge. User-tweet edges have been added between each user and their 200 tweets. Tweet-tweet edges have been added following the CLHG~\cite{wang2021cross} model. Each tweet is linked with the K most similar tweets according to cosine similarity (we set the value for K to 3). The cosine similarity was computed on the corresponding 768-dimensional DistilRoBERTa sentence embeddings.

The total number of nodes and edges for each dataset is summarized in Table~\ref{tab:num_nodes_edges}.

\begin{table}[]
  \caption{Number of nodes and edges in the created heterogeneous graphs for each of the four datasets.}
  \label{tab:num_nodes_edges}
  \begin{tabular}{r|c|c|c|c|c}
    \multicolumn{1}{l|}{} & \textbf{FN} & \textbf{HS} & \textbf{IS} & \textbf{SS} & \textbf{Y} \\ \hline
    \textbf{User nodes} & 500 & 200 & 420 & 140 & 883 \\
    \textbf{Tweet nodes} & 50,000 & 40,000 & 84,000 & 28,000 & 68,172 \\
    \textbf{Word nodes} & 3,506 & 2,713 & 8,580 & 3,394 & 5,557 \\
    \textbf{Total} & 54,006 & 42,913 & 93,000 & 31,534 & 74,612 \\ \hline
    \textbf{User-tweet edges} & 50,000 & 40,000 & 84,000 & 28,000 & 68,172 \\
    \textbf{Tweet-tweet edges} & 150,000 & 120,000 & 252,000 & 84,000 & 204,516 \\
    \textbf{Tweet-word edges} & 454,244 & 326,363 & 1,563,131 & 409,909 & 1,807,498 \\
    \textbf{Word-word edges} & 278,668 & 187,540 & 1,020,308 & 263,862 & 592,033 \\
    \textbf{Total} & 932,912 & 673,903 & 2,919,439 & 785,771 & 2,672,219
  \end{tabular}
\end{table}

\section{Graph Neural Network Models}
\label{sec:gnn_models}

In this research, three GNN architectures have been investigated for antisocial behavior detection: GraphSAGE~\cite{hamilton2017inductive}, Graph Attention Network (GAT)~\cite{velickovic2018graph}, and Graph Transformer~\cite{shi2020masked, dwivedi2021generalization}. GraphSAGE is an inductive methodology for graph representation learning using sampling and aggregation of features from a node’s local fixed-size neighborhood. Different tasks and problems are likely to leverage different aggregation functions (e.g., mean, LSTM pooling) and/or loss functions. GAT leverages masked self-attention layers in graph neural networks. The hidden representation of the nodes is computed with a self-attention mechanism that enables the nodes to attend to neighborhood features by specifying different weights for each neighbor node. Graph Transformer~\cite{shi2020masked} is a generalization of the Transformer architectures for graph structures. The attention mechanism is represented as a function of the neighborhood connectivity for each node in the graph and the positional encoding is represented by the Laplacian eigenvectors. The normalization layer is replaced by a batch normalization layer. The architecture could be extended to edge feature representation. Unified Message Passing (UniMP)~\cite{dwivedi2021generalization} jointly performs feature and label propagation by embedding the partially observed labels into the same space as node features. It is trained with a masked label prediction strategy inspired by BERT. We have used the modified Graph Transformer operator from the UniMP.

\begin{figure}
  \centering
\includegraphics[width=0.99\linewidth]{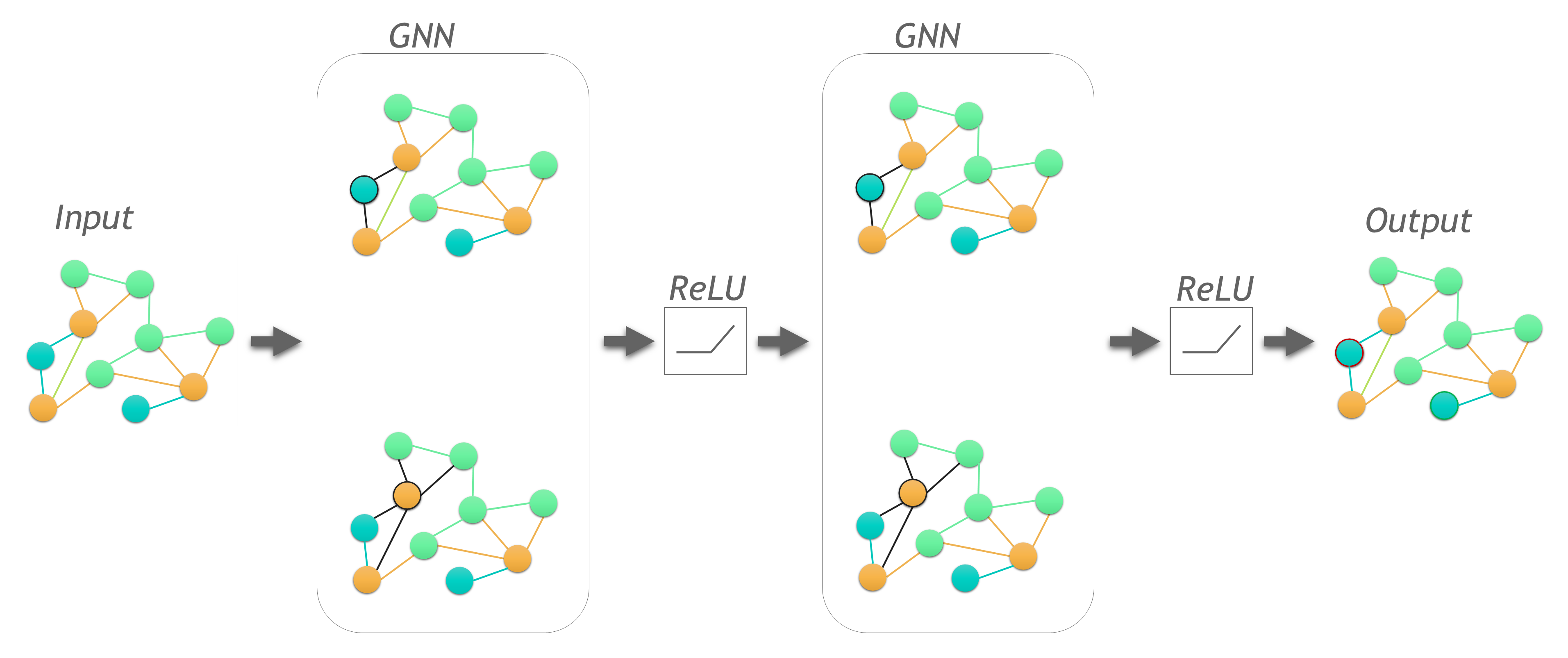}
  \caption{Architecture of a heterogeneous GNN model.}
  \label{fig:architecture}
\end{figure}

Our architecture is composed of a two-layer heterogeneous graph neural network followed by a ReLU activation that maps the nodes into a low-dimensional latent space. For the purpose of classifying nodes, a fully-connected layer has been added on top of the GNN model, which infers the class for the user nodes. The architecture is the same for all three models and is displayed in Figure~\ref{fig:architecture}.

We have used PyTorch implementation of these models available in the PyTorch Geometric library\footnote{https://pytorch-geometric.readthedocs.io/en/latest/, last visited: 25.02.2023}. The models have been created with the implementation for homogeneous graphs, and then are transformed into models suitable for heterogeneous graphs. All models have been trained with AdamW optimizer and binary cross-entropy loss. For the other hyperparameters, we have performed a hyperparameter search among a set of possible values. The optimal hyperparameters for each model and each dataset are summarized in Table~\ref{tab:gnns_hyperparameters}.

\begin{table}[]
  \caption{Optimal hyperparameters for GNN models.}
  \label{tab:gnns_hyperparameters}
  \begin{tabular}{r|c|c|c}
    \hline
    \multicolumn{4}{c}{\textbf{Fake News}} \\ \hline
    & \textbf{Learning Rate} & \textbf{Weight Decay} & \textbf{Epochs} \\ \hline
    \textbf{GraphSAGE} & 0.01 & 0.00005 & 250 \\
    \textbf{GAT} & 0.001 & 0.0005 & 250 \\
    \textbf{GraphTransformer} & 0.01 & 0.00005 & 500 \\ \hline \hline
    \multicolumn{4}{c}{\textbf{Hate Speech}} \\ \hline
    & \textbf{Learning Rate} & \textbf{Weight Decay} & \textbf{Epochs} \\ \hline
    \textbf{GraphSAGE} & 0.01 & 0.0005 & 50 \\
    \textbf{GAT} & 0.01 & 0.005 & 250 \\
    \textbf{GraphTransformer} & 0.001 & 0.00005 & 50 \\ \hline \hline
    \multicolumn{4}{c}{\textbf{Irony Stereotype}} \\ \hline
    & \textbf{Learning Rate} & \textbf{Weight Decay} & \textbf{Epochs} \\ \hline
    \textbf{GraphSAGE} & 0.01 & 0.05 & 50 \\
    \textbf{GAT} & 0.0001 & 0.00005 & 250 \\
    \textbf{GraphTransformer} & 0.001 & 0.05 & 250 \\ \hline \hline
    \multicolumn{4}{c}{\textbf{Stereotype Stance}} \\ \hline
    & \textbf{Learning Rate} & \textbf{Weight Decay} & \textbf{Epochs} \\ \hline
    \textbf{GraphSAGE} & 0.01 & 0.05 & 50 \\
    \textbf{GAT} & 0.01 & 0.005 & 50 \\
    \textbf{GraphTransformer} & 0.01 & 0.05 & 250 \\
    \hline \hline
    \multicolumn{4}{c}{\textbf{Yelp}} \\ \hline
    & \textbf{Learning Rate} & \textbf{Weight Decay} & \textbf{Epochs} \\ \hline
    \textbf{GraphSAGE} & 0.01 & 0.0005 & 100 \\
    \textbf{GAT} & 0.0001 & 0.0005 & 500 \\
    \textbf{GraphTransformer} & 0.0001 & 0.05 & 100 \\
    \hline
  \end{tabular}
\end{table}

\section{Results}
\label{sec:results}

We have performed several experiments with baseline Transformer models and GNN models. For each dataset, we have trained six models with the corresponding optimal hyperparameters shown in Table~\ref{tab:transformers_hyperparameters} and Table~\ref{tab:gnns_hyperparameters}. Each of the models has been trained on Quadro RTX 8000 48GB GPU.

\subsection{Comparison with Baseline Models}

To evaluate the models, accuracy has been calculated for the samples in the corresponding test sets. The results are shown in Table~\ref{tab:results}. Evaluation results of the three best performing models in the shared tasks are also included. For the fake news dataset, the test set provided by PAN has been used. For the other datasets, 10\% of the training set has been utilized for testing. The subtask of stereotype stance detection has been evaluated with the F1 measure. The results of the best performing models on this dataset are not shown since our models were not evaluated with the F1 measure.

The results show that for most of the cases, GNN models, in general, perform worse than the baseline Transformer models. Since deep neural networks require huge amounts of data for training and given that these datasets are relatively small, we could hypothesize that the worse performance is due to the small amount of data. On the other hand, Transformer-based models are pre-trained on large datasets which gives them a significant advantage over the other models.

For the hate speech dataset, both GraphSAGE and GraphTransformer models have the same accuracy as DistilBERT which is the second best model for the dataset. For the stereotype stance dataset, the GraphTransformer model has the same accuracy as RoBERTa which is the third best model for the dataset. The difference from the best performing model is 0.1 for the hate speech dataset and 0.14 for the stereotype stance dataset. These results demonstrate the capability of GNN models to successfully learn from graphs created from text data.

To compare with the best performing models in the PAN shared tasks, all three GNN models outperform the three best performing models for the hate speech dataset. However, for the fake news and irony stereotype datasets, the performance is inferior. The hate speech dataset is the smallest one among the three. Taking into account the fact that the models were not pre-trained, we could hypothesize that learning from a smaller graph is easier when the models are not pre-trained. Transformer-based models outperform the baseline models for the fake news and hate speech datasets. The DistilRoBERTa model has the best performance on the fake news dataset, while the RoBERTa model is the best performing model on the hate speech dataset. Nevertheless, we should point out that for all the datasets, except the fake news dataset, the evaluation was not done using the same test set, and therefore we could not know precisely how they would perform if the original test set was used.

\begin{table}[]
  \caption{Evaluation results and comparison with baseline models. The metric shown is accuracy. For the Stereotype Stance dataset participants were ranked according to the F1 measure and the results are not shown here.}
  \label{tab:results}
  \begin{tabular}{r|c|c|c|c|c}
    \hline
    & \textbf{\shortstack{Fake \\ News}} & \textbf{\shortstack{Hate \\ Speech}} & \textbf{\shortstack{Irony \\ Stereotype}} & \textbf{\shortstack{Stereotype \\ Stance}} & \textbf{\shortstack{Yelp \\ \textcolor{white}{.}}} \\ \hline
    \textbf{DistilBERT} & 0.72 & 0.80 & 0.83 & 0.93 & 0.74 \\
    \textbf{RoBERTa} & 0.72 & 0.90 & 0.90 & 0.79 & 0.78 \\
    \textbf{DistilRoBERTa} & 0.80 & 0.70 & 0.83 & 0.86 & 0.73 \\ \hline
    \textbf{GraphSAGE} & 0.54 & 0.80 & 0.60 & 0.71 & 0.63 \\
    \textbf{GAT} & 0.56 & 0.75 & 0.74 & 0.71 & 0.62 \\
    \textbf{GraphTransformer} & 0.55 & 0.80 & 0.67 & 0.79 & 0.64 \\ \hline
    \textbf{\#1} & 0.75 & 0.74 & 0.99 & / & / \\
    \textbf{\#2} & 0.75 & 0.73 & 0.98 & / & / \\
    \textbf{\#3} & 0.74 & 0.72 & 0.97 & / & / \\ \hline
  \end{tabular}
\end{table}

\subsection{Ablation Study}

To analyze the effectiveness of each component in the graph, ablation studies have been performed. Four variants of the heterogeneous graph have been examined:
\begin{itemize}
    \item \textit{all} - all components are included.
    \item \textit{no-word-word} - edges between word nodes are excluded from the graph. 
    \item \textit{no-word} - word nodes and edges that they are part of are excluded from the graph.
    \item \textit{no-doc-doc} - edges between tweet nodes are excluded from the graph.
\end{itemize}
A separate model has been trained using the optimal hyperparameters for each variant and accuracy on the test set was calculated. The results are summarized in Table~\ref{tab:ablation_results}.

The results show that the best performance is achieved when all the components in the graph are included. One exception is the GraphSAGE model on the irony stereotype dataset for which the best performance is achieved by the \textit{no-word} variant suggesting that removing word nodes and edges that they are part of leads to better performance than including all the components in the graph. This dataset is significantly bigger than the others and we can conclude that user and tweet nodes, as well as edges between them, are sufficient for the GraphSAGE model to successfully learn to classify the users. For the stereotype stance dataset, the GraphSAGE model achieved the same performances for all variants. The worst performance for all datasets is achieved with the \textit{no-doc-doc} variant indicating that removing the edges between tweet pairs reduces the performance. Edges between tweet pairs add shortcuts in the processing that could lead to faster convergence of the models and we could expect worse performances with their removal. An exception is the GraphTransformer model on the Yelp dataset for which the \textit{no-doc-doc} variant achieves the second best result. The structure of the reviews in the Yelp dataset differs from the tweets in the other Twitter datasets. 

For the GraphSAGE model, the \textit{no-word} variant is better than the \textit{no-word-word} variant for all Twitter datasets suggesting that removing any component that is related to words is better than removing only edges between word pairs. For the Yelp dataset, the \textit{no-word-word} variant is better. We could hypothesize that removing only edges between word pairs leads to better results for the Yelp reviews rather then removing any word related component. GraphTransformer follows the same pattern except for the stereotype stance dataset for which the \textit{no-word-word} variant achieved better results. GAT model has better results with the \textit{no-word-word} variant for the fake news and irony stereotype datasets, while the \textit{no-word} variant is better for the hate speech and stereotype stance datasets. The latter are smaller datasets and we could hypothesize that the GAT model could learn to better classify user nodes without any component related to words for smaller datasets.

\begin{table}[]
  \caption{Ablation results for the GNN models. The shown values represent the accuracy  metric on the test set.}
  \label{tab:ablation_results}
  \begin{tabular}{r|c|c|c|c}
    \hline
    \multicolumn{5}{c}{\textbf{Fake News}} \\ \hline
    & \textbf{all} & \textbf{no-word-word} & \textbf{no-word} & \textbf{no-doc-doc} \\ \hline
    \textbf{GraphSAGE} & \textbf{0.54} & 0.52 & 0.54 & 0.51 \\
    \textbf{GAT} & \textbf{0.56} & 0.51 & 0.50 & 0.47 \\
    \textbf{GraphTransformer} & \textbf{0.55} & 0.46 & 0.54 & 0.44 \\ \hline \hline
    \multicolumn{5}{c}{\textbf{Hate Speech}} \\ \hline
    & \textbf{all} & \textbf{no-word-word} & \textbf{no-word} & \textbf{no-doc-doc} \\ \hline
    \textbf{GraphSAGE} & \textbf{0.80} & 0.65 & 0.75 & 0.65 \\
    \textbf{GAT} & \textbf{0.75} & 0.60 & 0.70 & 0.65 \\
    \textbf{GraphTransformer} & \textbf{0.80} & 0.70 & 0.70 & 0.70 \\ \hline \hline
    \multicolumn{5}{c}{\textbf{Irony Stereotype}} \\ \hline
    & \textbf{all} & \textbf{no-word-word} & \textbf{no-word} & \textbf{no-doc-doc} \\ \hline
    \textbf{GraphSAGE} & 0.60 & 0.57 & \textbf{0.62} & 0.57 \\
    \textbf{GAT} & \textbf{0.74} & 0.55 & 0.50 & 0.50 \\
    \textbf{GraphTransformer} & \textbf{0.67} & 0.60 & 0.64 & 0.57 \\ \hline \hline
    \multicolumn{5}{c}{\textbf{Stereotype Stance}} \\ \hline
    & \textbf{all} & \textbf{no-word-word} & \textbf{no-word} & \textbf{no-doc-doc} \\ \hline
    \textbf{GraphSAGE} & \textbf{0.71} & 0.71 & 0.71 & 0.71 \\
    \textbf{GAT} & \textbf{0.71} & 0.50 & 0.71 & 0.71 \\
    \textbf{GraphTransformer} & \textbf{0.79} & 0.79 & 0.57 & 0.57 \\ \hline \hline
    \multicolumn{5}{c}{\textbf{Yelp}} \\ \hline
    & \textbf{all} & \textbf{no-word-word} & \textbf{no-word} & \textbf{no-doc-doc} \\ \hline
    \textbf{GraphSAGE} & \textbf{0.63} & 0.56 & 0.55 & 0.42 \\
    \textbf{GAT} & \textbf{0.62} & 0.55 & 0.53 & 0.47 \\
    \textbf{GraphTransformer} & \textbf{0.64} & 0.49 & 0.49 & 0.56 \\
    \hline
  \end{tabular}
\end{table}

\section{Conclusion}
\label{sec:conclusion}

This paper explored the performances of graph neural networks for the task of antisocial behavior detection on Twitter. Three GNN architectures (GraphSAGE, GAT, and Graph Transformer) were evaluated against four datasets composed of Twitter users and tweets that they have posted that were provided by PAN shared tasks, and one dataset composed of Yelp users and reviews that they have written that was extracted from the Yelp Open Dataset. A heterogeneous graph dataset has been created with user, tweet/review, and word nodes, as well as five types of edges between them.

An ablation study was performed to investigate which components of the heterogeneous graph have contributed the most. The results showed that the best performances are achieved when all the graph components are included, while the worst performances were obtained when the edges between tweet pairs were excluded from the graph.

Transformer-based models were also trained on the same datasets as baseline models for comparison. When compared against the baseline models, the GNN models showed inferior performance for most of the experiments. For the experiments, pre-trained Transformer-based models (DistilBERT, RoBERTa, and DistilRoBERTa) have been used. The models are pre-trained on large datasets which gives them a significant advantage. This hypothesis leads to a possible future direction which is to first pre-train GNNs on a larger dataset, and then train on the specific datasets that were used in this research.

For two of the datasets employed in this study, GNN models showed comparable performances with second best and third best Transformer-based models. These findings indicate the capability of GNN models to learn from the types of data derived from social networks that were utilized in this research. We anticipate that GNN models could be successfully applied to other text classification or even wider natural language processing or generation tasks.

\bibliography{sn-bibliography}


\begin{thebibliography}{23}
\ifx \bisbn   \undefined \def \bisbn  #1{ISBN #1}\fi
\ifx \binits  \undefined \def \binits#1{#1}\fi
\ifx \bauthor  \undefined \def \bauthor#1{#1}\fi
\ifx \batitle  \undefined \def \batitle#1{#1}\fi
\ifx \bjtitle  \undefined \def \bjtitle#1{#1}\fi
\ifx \bvolume  \undefined \def \bvolume#1{\textbf{#1}}\fi
\ifx \byear  \undefined \def \byear#1{#1}\fi
\ifx \bissue  \undefined \def \bissue#1{#1}\fi
\ifx \bfpage  \undefined \def \bfpage#1{#1}\fi
\ifx \blpage  \undefined \def \blpage #1{#1}\fi
\ifx \burl  \undefined \def \burl#1{\textsf{#1}}\fi
\ifx \doiurl  \undefined \def \doiurl#1{\url{https://doi.org/#1}}\fi
\ifx \betal  \undefined \def \betal{\textit{et al.}}\fi
\ifx \binstitute  \undefined \def \binstitute#1{#1}\fi
\ifx \binstitutionaled  \undefined \def \binstitutionaled#1{#1}\fi
\ifx \bctitle  \undefined \def \bctitle#1{#1}\fi
\ifx \beditor  \undefined \def \beditor#1{#1}\fi
\ifx \bpublisher  \undefined \def \bpublisher#1{#1}\fi
\ifx \bbtitle  \undefined \def \bbtitle#1{#1}\fi
\ifx \bedition  \undefined \def \bedition#1{#1}\fi
\ifx \bseriesno  \undefined \def \bseriesno#1{#1}\fi
\ifx \blocation  \undefined \def \blocation#1{#1}\fi
\ifx \bsertitle  \undefined \def \bsertitle#1{#1}\fi
\ifx \bsnm \undefined \def \bsnm#1{#1}\fi
\ifx \bsuffix \undefined \def \bsuffix#1{#1}\fi
\ifx \bparticle \undefined \def \bparticle#1{#1}\fi
\ifx \barticle \undefined \def \barticle#1{#1}\fi
\bibcommenthead
\ifx \bconfdate \undefined \def \bconfdate #1{#1}\fi
\ifx \botherref \undefined \def \botherref #1{#1}\fi
\ifx \url \undefined \def \url#1{\textsf{#1}}\fi
\ifx \bchapter \undefined \def \bchapter#1{#1}\fi
\ifx \bbook \undefined \def \bbook#1{#1}\fi
\ifx \bcomment \undefined \def \bcomment#1{#1}\fi
\ifx \oauthor \undefined \def \oauthor#1{#1}\fi
\ifx \citeauthoryear \undefined \def \citeauthoryear#1{#1}\fi
\ifx \endbibitem  \undefined \def \endbibitem {}\fi
\ifx \bconflocation  \undefined \def \bconflocation#1{#1}\fi
\ifx \arxivurl  \undefined \def \arxivurl#1{\textsf{#1}}\fi
\csname PreBibitemsHook\endcsname

%%% 1
\bibitem[\protect\citeauthoryear{Kipf and Welling}{2017}]{kipf2017semi}
\begin{botherref}
\oauthor{\bsnm{Kipf}, \binits{T.N.}},
\oauthor{\bsnm{Welling}, \binits{M.}}:
Semi-supervised classification with graph convolutional networks.
5th International Conference on Learning Representations, {ICLR} 2017, Toulon,
  France, April 24-26, 2017, Conference Track Proceedings
(2017)
\end{botherref}
\endbibitem

%%% 2
\bibitem[\protect\citeauthoryear{Defferrard
  et~al.}{2016}]{defferrard2016convolutional}
\begin{botherref}
\oauthor{\bsnm{Defferrard}, \binits{M.}},
\oauthor{\bsnm{Bresson}, \binits{X.}},
\oauthor{\bsnm{Vandergheynst}, \binits{P.}}:
Convolutional neural networks on graphs with fast localized spectral filtering.
Advances in neural information processing systems
\textbf{29}
(2016)
\end{botherref}
\endbibitem

%%% 3
\bibitem[\protect\citeauthoryear{Hamilton et~al.}{2017}]{hamilton2017inductive}
\begin{bchapter}
\bauthor{\bsnm{Hamilton}, \binits{W.L.}},
\bauthor{\bsnm{Ying}, \binits{R.}},
\bauthor{\bsnm{Leskovec}, \binits{J.}}:
\bctitle{Inductive representation learning on large graphs}.
In: \bbtitle{Proceedings of the 31st International Conference on Neural
  Information Processing Systems},
pp. \bfpage{1025}--\blpage{1035}
(\byear{2017})
\end{bchapter}
\endbibitem

%%% 4
\bibitem[\protect\citeauthoryear{Wu et~al.}{2021}]{wu2021graph}
\begin{botherref}
\oauthor{\bsnm{Wu}, \binits{L.}},
\oauthor{\bsnm{Chen}, \binits{Y.}},
\oauthor{\bsnm{Shen}, \binits{K.}},
\oauthor{\bsnm{Guo}, \binits{X.}},
\oauthor{\bsnm{Gao}, \binits{H.}},
\oauthor{\bsnm{Li}, \binits{S.}},
\oauthor{\bsnm{Pei}, \binits{J.}},
\oauthor{\bsnm{Long}, \binits{B.}}:
Graph neural networks for natural language processing: A survey.
CoRR
(2021)
\end{botherref}
\endbibitem

%%% 5
\bibitem[\protect\citeauthoryear{Yao et~al.}{2019}]{yao2019graph}
\begin{bchapter}
\bauthor{\bsnm{Yao}, \binits{L.}},
\bauthor{\bsnm{Mao}, \binits{C.}},
\bauthor{\bsnm{Luo}, \binits{Y.}}:
\bctitle{Graph convolutional networks for text classification}.
In: \bbtitle{Proceedings of the AAAI Conference on Artificial Intelligence},
vol. \bseriesno{33},
pp. \bfpage{7370}--\blpage{7377}
(\byear{2019})
\end{bchapter}
\endbibitem

%%% 6
\bibitem[\protect\citeauthoryear{Lu et~al.}{2020}]{lu2020vgcn}
\begin{bchapter}
\bauthor{\bsnm{Lu}, \binits{Z.}},
\bauthor{\bsnm{Du}, \binits{P.}},
\bauthor{\bsnm{Nie}, \binits{J.-Y.}}:
\bctitle{Vgcn-bert: augmenting bert with graph embedding for text
  classification}.
In: \bbtitle{European Conference on Information Retrieval},
pp. \bfpage{369}--\blpage{382}
(\byear{2020}).
\bcomment{Springer}
\end{bchapter}
\endbibitem

%%% 7
\bibitem[\protect\citeauthoryear{Lin et~al.}{2021}]{lin2021bertgcn}
\begin{bchapter}
\bauthor{\bsnm{Lin}, \binits{Y.}},
\bauthor{\bsnm{Meng}, \binits{Y.}},
\bauthor{\bsnm{Sun}, \binits{X.}},
\bauthor{\bsnm{Han}, \binits{Q.}},
\bauthor{\bsnm{Kuang}, \binits{K.}},
\bauthor{\bsnm{Li}, \binits{J.}},
\bauthor{\bsnm{Wu}, \binits{F.}}:
\bctitle{Bertgcn: Transductive text classification by combining gnn and bert}.
In: \bbtitle{Findings of the Association for Computational Linguistics:
  ACL-IJCNLP 2021},
pp. \bfpage{1456}--\blpage{1462}
(\byear{2021})
\end{bchapter}
\endbibitem

%%% 8
\bibitem[\protect\citeauthoryear{Rangel et~al.}{2020}]{rangel2020overview}
\begin{bchapter}
\bauthor{\bsnm{Rangel}, \binits{F.}},
\bauthor{\bsnm{Giachanou}, \binits{A.}},
\bauthor{\bsnm{Ghanem}, \binits{B.H.H.}},
\bauthor{\bsnm{Rosso}, \binits{P.}}:
\bctitle{Overview of the 8th author profiling task at pan 2020: Profiling fake
  news spreaders on twitter}.
In: \bbtitle{CEUR Workshop Proceedings},
vol. \bseriesno{2696},
pp. \bfpage{1}--\blpage{18}
(\byear{2020}).
\bcomment{Sun SITE Central Europe}
\end{bchapter}
\endbibitem

%%% 9
\bibitem[\protect\citeauthoryear{Rangel et~al.}{2021}]{rangel2021profiling}
\begin{bchapter}
\bauthor{\bsnm{Rangel}, \binits{F.}},
\bauthor{\bsnm{Pe{\~n}a-Sarrac{\'e}n}, \binits{G.L.d.l.}},
\bauthor{\bsnm{Chulvi-Ferriols}, \binits{M.A.}},
\bauthor{\bsnm{Fersini}, \binits{E.}},
\bauthor{\bsnm{Rosso}, \binits{P.}}:
\bctitle{Profiling hate speech spreaders on twitter task at pan 2021}.
In: \bbtitle{Proceedings of the Working Notes of CLEF 2021, Conference and Labs
  of the Evaluation Forum, Bucharest, Romania, September 21st to 24th, 2021},
pp. \bfpage{1772}--\blpage{1789}
(\byear{2021}).
\bcomment{CEUR}
\end{bchapter}
\endbibitem

%%% 10
\bibitem[\protect\citeauthoryear{Reynier et~al.}{2022}]{reynier2022profiling}
\begin{botherref}
\oauthor{\bsnm{Reynier}, \binits{O.-B.}},
\oauthor{\bsnm{Berta}, \binits{C.}},
\oauthor{\bsnm{Francisco}, \binits{R.}},
\oauthor{\bsnm{Paolo}, \binits{R.}},
\oauthor{\bsnm{Elisabetta}, \binits{F.}}:
Profiling irony and stereotype spreaders on twitter (irostereo) at pan 2022.
CEUR-WS. org
(2022)
\end{botherref}
\endbibitem

%%% 11
\bibitem[\protect\citeauthoryear{Buda and Bolonyai}{2020}]{buda2020ensemble}
\begin{bchapter}
\bauthor{\bsnm{Buda}, \binits{J.}},
\bauthor{\bsnm{Bolonyai}, \binits{F.}}:
\bctitle{{An Ensemble Model Using N-grams and Statistical Featuresto Identify
  Fake News Spreaders on Twitter---Notebook for PAN at CLEF 2020}}.
(\byear{2020})
\end{bchapter}
\endbibitem

%%% 12
\bibitem[\protect\citeauthoryear{Siino et~al.}{2021}]{siino2021detection}
\begin{bchapter}
\bauthor{\bsnm{Siino}, \binits{M.}},
\bauthor{\bsnm{{Di Nuovo}}, \binits{E.}},
\bauthor{\bsnm{Tinnirello}, \binits{I.}},
\bauthor{\bsnm{{La Cascia}}, \binits{M.}}:
\bctitle{{Detection of hate speech spreaders using convolutional neural
  networks---Notebook for PAN at CLEF 2021}}.
(\byear{2021})
\end{bchapter}
\endbibitem

%%% 13
\bibitem[\protect\citeauthoryear{Yu et~al.}{2022}]{yu2022bert}
\begin{bchapter}
\bauthor{\bsnm{Yu}, \binits{W.}},
\bauthor{\bsnm{Boenninghoff}, \binits{B.}},
\bauthor{\bsnm{Kolossa}, \binits{D.}}:
\bctitle{{BERT-based ironic authors profiling}}.
(\byear{2022})
\end{bchapter}
\endbibitem

%%% 14
\bibitem[\protect\citeauthoryear{Sanh et~al.}{2019}]{sanh2019distilbert}
\begin{botherref}
\oauthor{\bsnm{Sanh}, \binits{V.}},
\oauthor{\bsnm{Debut}, \binits{L.}},
\oauthor{\bsnm{Chaumond}, \binits{J.}},
\oauthor{\bsnm{Wolf}, \binits{T.}}:
Distilbert, a distilled version of bert: smaller, faster, cheaper and lighter.
arXiv preprint arXiv:1910.01108
(2019)
\end{botherref}
\endbibitem

%%% 15
\bibitem[\protect\citeauthoryear{Liu et~al.}{2019}]{liu2019roberta}
\begin{botherref}
\oauthor{\bsnm{Liu}, \binits{Y.}},
\oauthor{\bsnm{Ott}, \binits{M.}},
\oauthor{\bsnm{Goyal}, \binits{N.}},
\oauthor{\bsnm{Du}, \binits{J.}},
\oauthor{\bsnm{Joshi}, \binits{M.}},
\oauthor{\bsnm{Chen}, \binits{D.}},
\oauthor{\bsnm{Levy}, \binits{O.}},
\oauthor{\bsnm{Lewis}, \binits{M.}},
\oauthor{\bsnm{Zettlemoyer}, \binits{L.}},
\oauthor{\bsnm{Stoyanov}, \binits{V.}}:
Roberta: A robustly optimized bert pretraining approach.
arXiv preprint arXiv:1907.11692
(2019)
\end{botherref}
\endbibitem

%%% 16
\bibitem[\protect\citeauthoryear{Sajjad et~al.}{2020}]{sajjad2020effect}
\begin{botherref}
\oauthor{\bsnm{Sajjad}, \binits{H.}},
\oauthor{\bsnm{Dalvi}, \binits{F.}},
\oauthor{\bsnm{Durrani}, \binits{N.}},
\oauthor{\bsnm{Nakov}, \binits{P.}}:
On the effect of dropping layers of pre-trained transformer models.
CoRR
(2020)
\end{botherref}
\endbibitem

%%% 17
\bibitem[\protect\citeauthoryear{Buciluǎ et~al.}{2006}]{buciluǎ2006model}
\begin{bchapter}
\bauthor{\bsnm{Buciluǎ}, \binits{C.}},
\bauthor{\bsnm{Caruana}, \binits{R.}},
\bauthor{\bsnm{Niculescu-Mizil}, \binits{A.}}:
\bctitle{Model compression}.
In: \bbtitle{Proceedings of the 12th ACM SIGKDD International Conference on
  Knowledge Discovery and Data Mining},
pp. \bfpage{535}--\blpage{541}
(\byear{2006})
\end{bchapter}
\endbibitem

%%% 18
\bibitem[\protect\citeauthoryear{Hinton et~al.}{2015}]{hinton2015distilling}
\begin{botherref}
\oauthor{\bsnm{Hinton}, \binits{G.}},
\oauthor{\bsnm{Vinyals}, \binits{O.}},
\oauthor{\bsnm{Dean}, \binits{J.}}:
Distilling the knowledge in a neural network.
arXiv preprint arXiv:1503.02531
(2015)
\end{botherref}
\endbibitem

%%% 19
\bibitem[\protect\citeauthoryear{Pennington et~al.}{2014}]{pennington2014glove}
\begin{bchapter}
\bauthor{\bsnm{Pennington}, \binits{J.}},
\bauthor{\bsnm{Socher}, \binits{R.}},
\bauthor{\bsnm{Manning}, \binits{C.D.}}:
\bctitle{Glove: Global vectors for word representation}.
In: \bbtitle{Proceedings of the 2014 Conference on Empirical Methods in Natural
  Language Processing (EMNLP)},
pp. \bfpage{1532}--\blpage{1543}
(\byear{2014})
\end{bchapter}
\endbibitem

%%% 20
\bibitem[\protect\citeauthoryear{Wang et~al.}{2021}]{wang2021cross}
\begin{bchapter}
\bauthor{\bsnm{Wang}, \binits{Z.}},
\bauthor{\bsnm{Liu}, \binits{X.}},
\bauthor{\bsnm{Yang}, \binits{P.}},
\bauthor{\bsnm{Liu}, \binits{S.}},
\bauthor{\bsnm{Wang}, \binits{Z.}}:
\bctitle{Cross-lingual text classification with heterogeneous graph neural
  network}.
In: \bbtitle{Proceedings of the 59th Annual Meeting of the Association for
  Computational Linguistics and the 11th International Joint Conference on
  Natural Language Processing (Volume 2: Short Papers)},
pp. \bfpage{612}--\blpage{620}
(\byear{2021})
\end{bchapter}
\endbibitem

%%% 21
\bibitem[\protect\citeauthoryear{Veli{\v{c}}kovi{\'{c}}
  et~al.}{2018}]{velickovic2018graph}
\begin{botherref}
\oauthor{\bsnm{Veli{\v{c}}kovi{\'{c}}}, \binits{P.}},
\oauthor{\bsnm{Cucurull}, \binits{G.}},
\oauthor{\bsnm{Casanova}, \binits{A.}},
\oauthor{\bsnm{Romero}, \binits{A.}},
\oauthor{\bsnm{Li{\`{o}}}, \binits{P.}},
\oauthor{\bsnm{Bengio}, \binits{Y.}}:
{Graph Attention Networks}.
International Conference on Learning Representations
(2018).
accepted as poster
\end{botherref}
\endbibitem

%%% 22
\bibitem[\protect\citeauthoryear{Shi et~al.}{2020}]{shi2020masked}
\begin{botherref}
\oauthor{\bsnm{Shi}, \binits{Y.}},
\oauthor{\bsnm{Huang}, \binits{Z.}},
\oauthor{\bsnm{Feng}, \binits{S.}},
\oauthor{\bsnm{Zhong}, \binits{H.}},
\oauthor{\bsnm{Wang}, \binits{W.}},
\oauthor{\bsnm{Sun}, \binits{Y.}}:
Masked label prediction: Unified message passing model for semi-supervised
  classification.
CoRR
(2020)
\end{botherref}
\endbibitem

%%% 23
\bibitem[\protect\citeauthoryear{Dwivedi and
  Bresson}{2021}]{dwivedi2021generalization}
\begin{botherref}
\oauthor{\bsnm{Dwivedi}, \binits{V.P.}},
\oauthor{\bsnm{Bresson}, \binits{X.}}:
A generalization of transformer networks to graphs.
AAAI Workshop on Deep Learning on Graphs: Methods and Applications
(2021)
\end{botherref}
\endbibitem

\end{thebibliography}

\end{document}